%% file: main.tex
\documentclass[10pt,twocolumn,letterpaper]{article}

\usepackage[pagenumbers]{cvpr}
\usepackage{tabularx}
\usepackage{multirow}
\usepackage[table]{xcolor}
\usepackage{cuted}
\usepackage{caption}

\newcommand{\methodname}{SuperNav-1.5}
\newcommand{\Methodname}{SuperNav-1.5}
\definecolor{cvprblue}{rgb}{0.21,0.49,0.74}
\usepackage[pagebackref,breaklinks,colorlinks,allcolors=cvprblue]{hyperref}

\def\confName{CVPR}
\def\confYear{2027}

\title{SuperNav-1.5: Structured Spatial Memory For Agentic Navigation Systems}

\author{
Jiarui Guo$^{1}$ \quad
Jinkai Zhang$^{1}$ \quad
Yuanhong Yu$^{1}$ \quad
Yongchen Zai$^{2}$ \\
Futang Zhu$^{2}$ \quad
Ruizhen Hu$^{3}$ \quad
Sida Peng$^{1}$ \quad
Xiaowei Zhou$^{1}$ \\[3pt]
$^{1}$Zhejiang University \quad
$^{2}$BYD New Technology Research Institute \quad
$^{3}$Shenzhen University \\
}

\begin{document}
\maketitle
\input{fig/teaser}
\input{sec/0_abstract}
\input{sec/introduction.tex}
\input{sec/related.tex}
\input{sec/method.tex}
\input{sec/experiments.tex}
\input{sec/conclusion.tex}

\section*{LLM Usage Statement}
\texttt{GPT-5.6-sol} was used as a development and writing assistant for code generation and revision, execution of experiment pipelines, and grammatical and sentence-level editing of the manuscript. All generated code, experimental procedures, and textual edits were reviewed and verified by the authors. The research ideas, analysis and interpretation of experimental results, initial manuscript draft, and scientific argument and organization were developed by the authors. Because this work studies agentic navigation systems, \texttt{GPT-5.6-terra}, \texttt{GPT-6.1-sol}, \texttt{KIMI-k3-256k}, and \texttt{GLM-5.3-flash} were additionally used as reasoning or policy models within the evaluated systems. Their outputs constitute experimental system behavior rather than assistance with research ideation or result interpretation.

{
    \small
    \bibliographystyle{ieeenat_fullname}
    \bibliography{main}
}

\input{sec/supplementary.tex}

\end{document}

%% file: fig/teaser.tex
\begin{strip}
  \centering
  \includegraphics[width=1.0\linewidth]{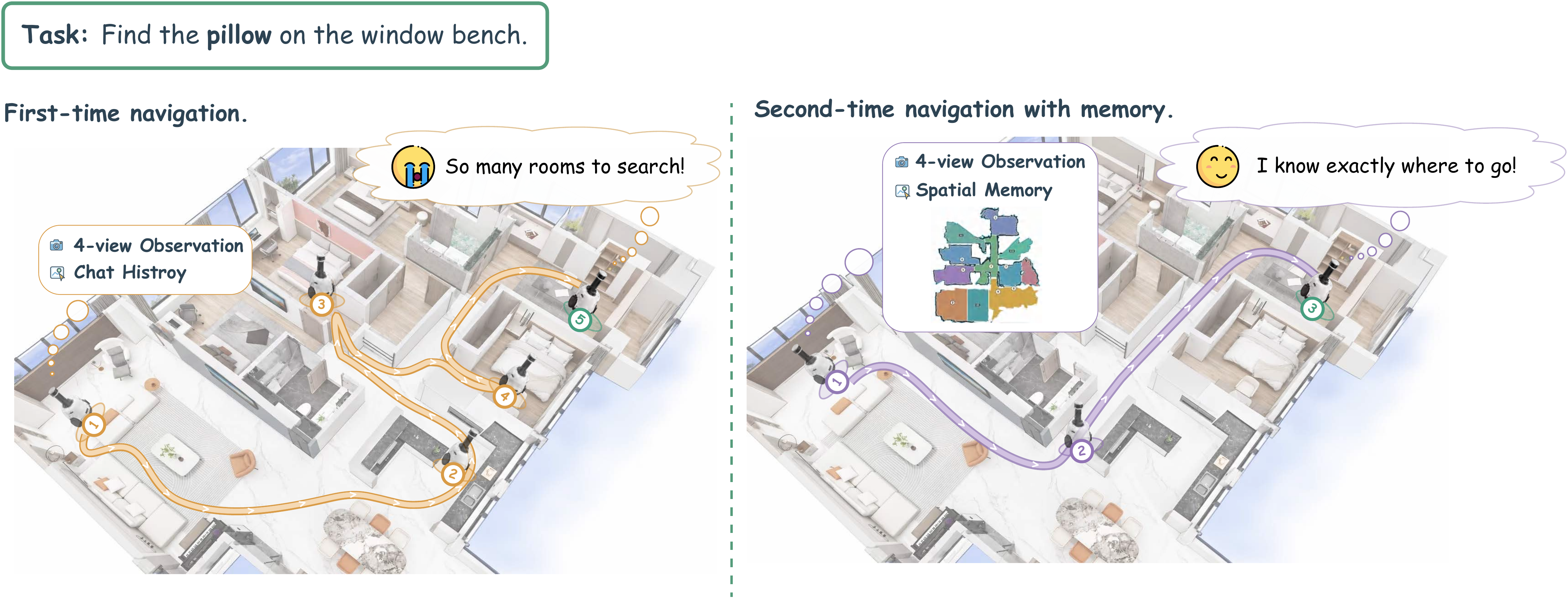}
  \captionsetup{hypcap=false}
  \captionof{figure}{\textbf{Navigation with Structured Spatial Memory.}
  In first time exploration, the agent repeatedly explores potential areas for target object. In second time navigation, the agent navigates with higher path efficiency using spatial memory.}
  \captionsetup{hypcap=true}
\end{strip}

%% file: sec/0_abstract.tex
\begin{abstract}
This paper addresses the challenge of scene-level memory design for efficient navigation across sessions and successive object goals.
Recent methods commonly represent a scene using scene graphs or image--text records.
However, dense scene-graph representations impose substantial reasoning and context overhead on multimodal large language models, while image--text records cannot be directly converted into executable geometric routes.
To address these limitations, we propose \textbf{\methodname}, an agentic navigation system equipped with a structured spatial memory constructed by the agent.
The memory consists of an area-segmented occupancy map annotated with agent-maintained, map-grounded memory nodes.
This map guides the agent toward unvisited areas and provides executable routes to known locations.
For multi-session navigation, we introduce agent-guided relocalization, whereby the agent infers an area prior from current observations and stored memory records to constrain geometric matching and localize the robot within the stored map.
Together, these designs enable efficient navigation across session and successive object goals.
Experiments show that our method outperforms baselines on single-object, multi-object, and demand-driven navigation benchmarks, with particularly strong gains in path efficiency.
Multi-session evaluation demonstrates that the constructed memory can be reused, while deployment on a real quadruped robot further shows the system's applicability. Project Page: \url{https://zju3dv.github.io/SuperNav-1.5/}
\end{abstract}

%% file: sec/introduction.tex
\section{Introduction}
\label{sec:introduction}

Reliable navigation requires understanding current observations and remembering the environment beyond the robot's immediate view.
For instance, within a single search, a robot should distinguish explored from unexplored areas to avoid redundant exploration~\cite{chaplot2020semexp}.
Across successive goals or repeated visits, it should reuse past experience to reach remembered destinations efficiently~\cite{chang2024goat}.
Together, these settings require persistent spatial memory that is both easy for an agent to query and geometrically grounded for navigation.

Existing approaches to spatial memory broadly fall into two categories: graph-based representations and recallable image--text records.
Graph-based spatial memories provide explicit structure but can be costly for an agent to process.
Scene graphs organize objects and rooms as nodes connected by relational edges~\cite{yin2024sgnav}, while observation-based topological memories retain visual features from past observations~\cite{li2024memonav,ge2026evomemnav}.
As exploration adds nodes and relations, exposing the full accumulated representation can increase context and reasoning costs, motivating selective memory access~\cite{ge2026evomemnav}.
The challenge is to expose useful spatial knowledge without overwhelming the agent with scene detail.

Recallable image--text records provide access to past visual observations and textual summary ~\cite{li2026agenticnav,liu2026navprobe}.
Although these systems retain spatial evidence, retrieving an image or textual record does not by itself specify a traversable route to a distant destination.
The remaining challenge is therefore to couple compact semantic recall with a geometric representation that can provide executable navigation route.

We introduce \textbf{\methodname}, an agentic navigation system with structured spatial memory that the agent selectively constructs and uses.
Our key insight is to use a geometric map as the spatial backbone while letting the agent decide where to retain useful memories.
During navigation, a real-time two-dimensional simultaneous localization and mapping (SLAM) system ~\cite{hess2016realtime} builds an occupancy map partitioned into areas.
The agent adaptively creates memory nodes at selected map locations, grounding retained scene information in a shared spatial reference.
This design simplifies memory access and management, while grounding memory nodes in the map enables efficient navigation to known areas.

This structured spatial memory supports both exploration and navigation to remembered locations.
During initial exploration, the occupancy map distinguishes explored from unexplored areas, guiding the agent toward unvisited regions.
To reach an unvisited area or revisit a remembered location, the agent selects the corresponding destination on the map. A map-based geometric tool then plans and executes the route, avoiding the need for the agent to infer step-by-step semantic instructions.

When reusing memory across navigation sessions, the robot must first determine its location within the stored map.
Our \emph{agent-guided relocalization} combines agent reasoning with geometric relocalization to estimate this location using limited geometric observations from the current session.
Once relocalized, the system reuses stored memory nodes and map geometry to navigate to subsequent goals.

Experiments evaluate \methodname{} on single-object, multi-object, demand-driven, and multi-session navigation.
On HM3D-OVON~\cite{yokoyama2024hm3dovon} val-unseen, our method achieves an 80.00\% success rate at the 1\,m evaluation threshold, outperforming the reported baselines.
On Habitat-GS~\cite{xia2026habitat} multi-object tasks, it improves success weighted by path length (SPL) from SuperNav's 0.1388 to 0.2494.
In second-session navigation from a different spawn position, memory reuse raises SPL from 0.4364 to 0.5022 compared with the no-memory setting.
These results support the value of geometry-grounded memory for long-horizon navigation and multi-session reuse.

Our contributions are threefold:
\begin{itemize}
    \item A structured spatial memory with agent-selected nodes grounded in a persistent two-dimensional occupancy map, linking scene recall to navigable locations.
    \item An agent-guided relocalization method that locates the robot in a stored map using limited geometric observations from the current session.
    \item A unified agentic navigation system integrating map-guided exploration, geometric navigation, and relocalization for long-horizon navigation and multi-session memory reuse.
\end{itemize}

%% file: sec/related.tex
\section{Related Work}
\label{sec:related}

\begin{figure*}[t!]
    \centering
    \includegraphics[width=\textwidth]{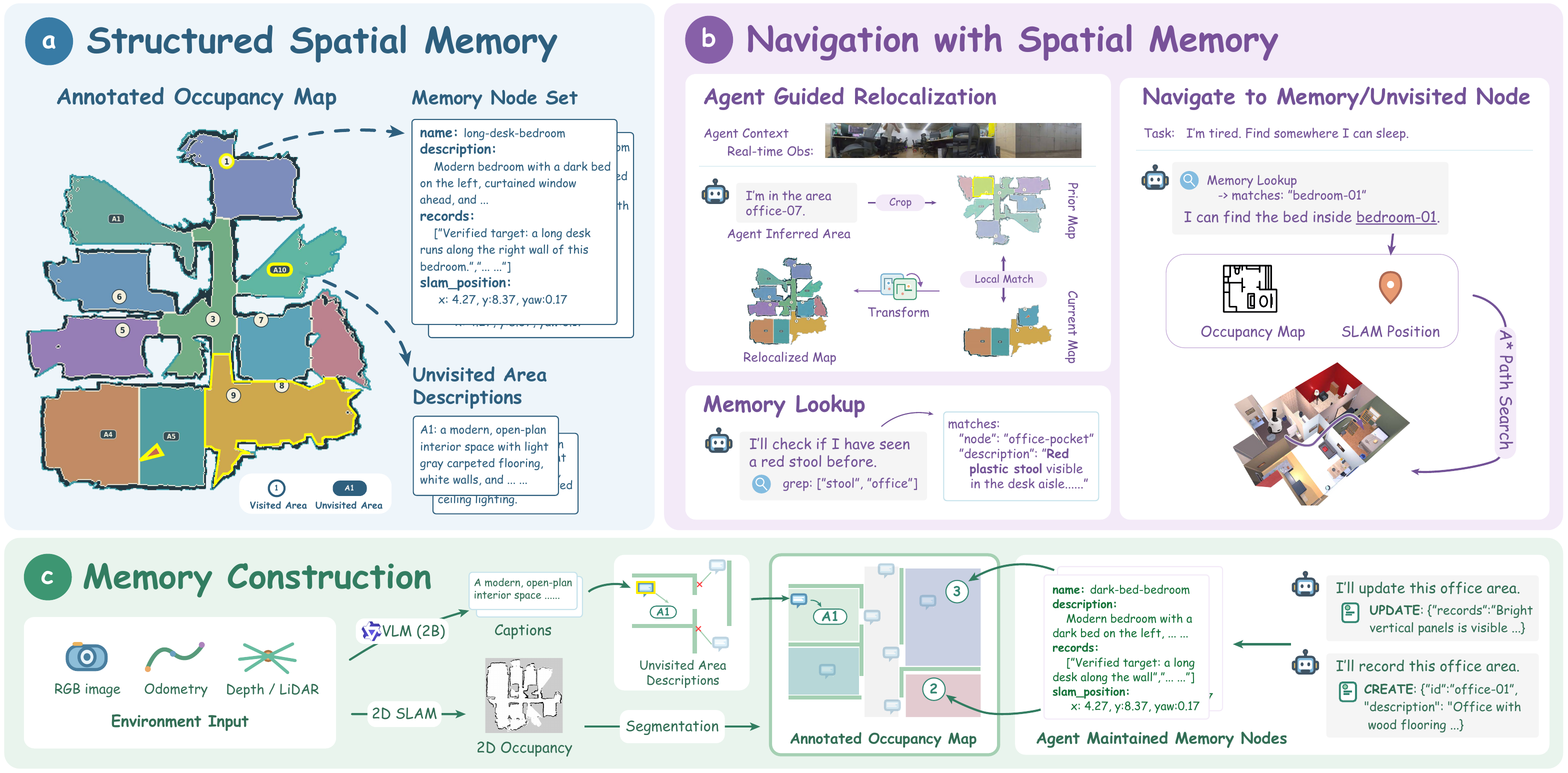}
    \caption{\textbf{Overview of \methodname.}
    (a) Structured spatial memory combines an area-segmented occupancy map with agent-maintained memory nodes and descriptions of unvisited areas.
    (b) The agent uses this memory through agent-guided relocalization, memory lookup, and geometric navigation to memory or unvisited-area locations.
    (c) Mapping and keyframe captioning run asynchronously, while the agent selectively creates and updates map-grounded memory nodes.}
    \label{fig:pipeline}
\end{figure*}

\subsection{Map-guided Navigation Systems}
Map-guided systems ground exploration decisions in spatial representations.
SemExp~\cite{chaplot2020semexp} and L3MVN~\cite{yu2023l3mvn} use semantic maps for learned or language-guided exploration. InstructNav~\cite{long2025instructnav} and VLFM~\cite{yokoyama2024vlfm} ground decisions in value maps.
ApexNav~\cite{zhang2025apexnav} combines adaptive semantic and geometric exploration with target-centric fusion, while VoroNav~\cite{wu2024voronav} selects waypoints on a reduced Voronoi graph.
However, exploration-oriented map states alone do not provide task-independent scene memory or support multi-session reuse.
We couple geometry with agent-selected memories and relocalization to support memory reuse beyond the current search.

\subsection{Agentic Navigation Systems}
Agentic navigation systems pair pretrained reasoning models with perception and navigation tools.
SuperNav~\cite{zhang2026supernav} provides navigation skills and a visual-point interface for the reasoning model, while DemandAgent~\cite{anonymous2026demandagent} introduces an embodied agent harness for interpreting high-level human demands.
AgenticNav~\cite{li2026agenticnav} and NavProbe~\cite{liu2026navprobe} expose selective visual or geometric evidence-retrieval tools to the agent.
Recent systems further incorporate persistent state and memory~\cite{chen2026harnessvln,zhao2026navharness}.
Although these systems establish flexible agent--environment interaction, an agentic interface alone does not determine how long-term spatial information is managed.
We therefore couple agentic tool use with structured, map-grounded memory and geometric navigation.

\subsection{Spatial Memory for Long-Horizon Navigation}
Spatial memory supports experience reuse across successive goals.
GOAT~\cite{chang2024goat}, SG-Nav~\cite{yin2024sgnav}, and SysNav~\cite{zhu2026sysnav} use instance-aware memory, scene graphs, or structured room-level representations.
MemoNav~\cite{li2024memonav} selects informative topological features, while EvoMemNav~\cite{ge2026evomemnav} combines per-visit visual nodes over an occupancy grid with budgeted retrieval and reflection-driven updates.
Memory is also central to recent agentic systems. AgenticNav~\cite{li2026agenticnav} and NavProbe~\cite{liu2026navprobe} retain visual or geometric evidence for later retrieval. HarnessVLN~\cite{chen2026harnessvln} maintains event memory and a spatiotemporal graph while NavHarness~\cite{zhao2026navharness} preserves maps and task knowledge across sessions.
Existing designs therefore span retrieval-oriented records and structured topological or scene graphs. The former do not necessarily provide executable geometric route towards destinations, whereas the latter require maintaining relational state.
In contrast, we represent spatial memory as map-grounded nodes without explicit graph edges and couple them with relocalization and geometric navigation, making selected memories directly actionable across sessions.

%% file: sec/method.tex
\section{Method}
\label{sec:method}

\label{sec:method-overview}
\Methodname{} equips the agentic navigation framework of SuperNav~\cite{zhang2026supernav} with persistent structured spatial memory for multi-object and multi-session navigation.
Given a navigation request and online observations, the agent uses this memory to select unexplored areas or recall previously visited destinations.
The key design is to let the occupancy map encode navigable geometry while the agent decides which scene information to retain, connecting semantic decisions to executable geometric goals.
As shown in \cref{fig:pipeline}, our system separates the memory representation (\cref{sec:spatial-memory}), its construction (\cref{sec:memory-construction}), and its use for navigation (\cref{sec:navigation-memory}).

\subsection{Structured Spatial Memory}
\label{sec:spatial-memory}
Our structured spatial memory is designed to make semantic information manageable for the agent while supporting executable geometric routes.
It is represented as an occupancy map annotated with agent-maintained, map-grounded memory nodes.
The annotated two-dimensional occupancy map provides a unified interface through which the agent accesses spatial memory (\cref{fig:pipeline}a).

The map displays color-coded areas, the current robot pose, and two types of markers: automatically generated unvisited-area entries and agent-maintained memory nodes.
The unvisited-area entries describe areas observed from nearby but not yet represented by an agent-created memory node, preserving incidental observations as leads for further exploration.
Each agent-maintained memory node contains a \emph{name}, a scene \emph{description}, factual \emph{records}, and a \emph{SLAM pose} $(x,y,\theta)$.
The node name supports direct lookup, the description and records preserve scene information, and the pose anchors that information to geometric navigation.
A dictionary maps each displayed marker identifier to its corresponding memory-node name or area entry.
Together, these annotations let the agent distinguish remembered locations from areas awaiting exploration and select destinations in a shared spatial reference.

\subsection{Memory Construction}
\label{sec:memory-construction}
The structured spatial memory consists of an underlying geometric map and area annotations. The geometric map is constructed asynchronously as the robot moves. The agent manages memory nodes through tool calls, whereas descriptions of unvisited areas are generated automatically for regions observed in passing but not yet explored.

\paragraph{Asynchronous geometric mapping.}
The occupancy map provides a continuously updated geometric foundation independent of the agent's memory decisions.
During navigation, Cartographer~\cite{hess2016realtime} asynchronously integrates depth or LiDAR observations with odometry to estimate the robot pose and construct an occupancy map.
Watershed segmentation~\cite{vincent1991watersheds} partitions the map into areas, which are assigned distinct colors for the agent-facing visualization (\cref{fig:pipeline}c).
By explicitly encoding spatial relationships, the resulting map frees the agent from managing them directly and guides it toward unvisited areas.
Further details of the geometric mapping procedure are provided in the supplementary material (\cref{sec:supp-occupancy}).

\paragraph{Agent-maintained memory nodes.}
The agent controls which observations become persistent memory through node-creation and update tools.
It can create a named node with a description and factual records, or revise an existing node as new information becomes available.
After either operation, the affected node is automatically bound to the robot's current SLAM pose.
This node-management mechanism allows the agent to selectively record important semantic information while preserving a geometric location for each node.
Further details of the memory-node design are provided in the supplementary material (\cref{sec:supp-memory-nodes}).

\begin{figure*}[t]
    \centering
    \includegraphics[width=\textwidth]{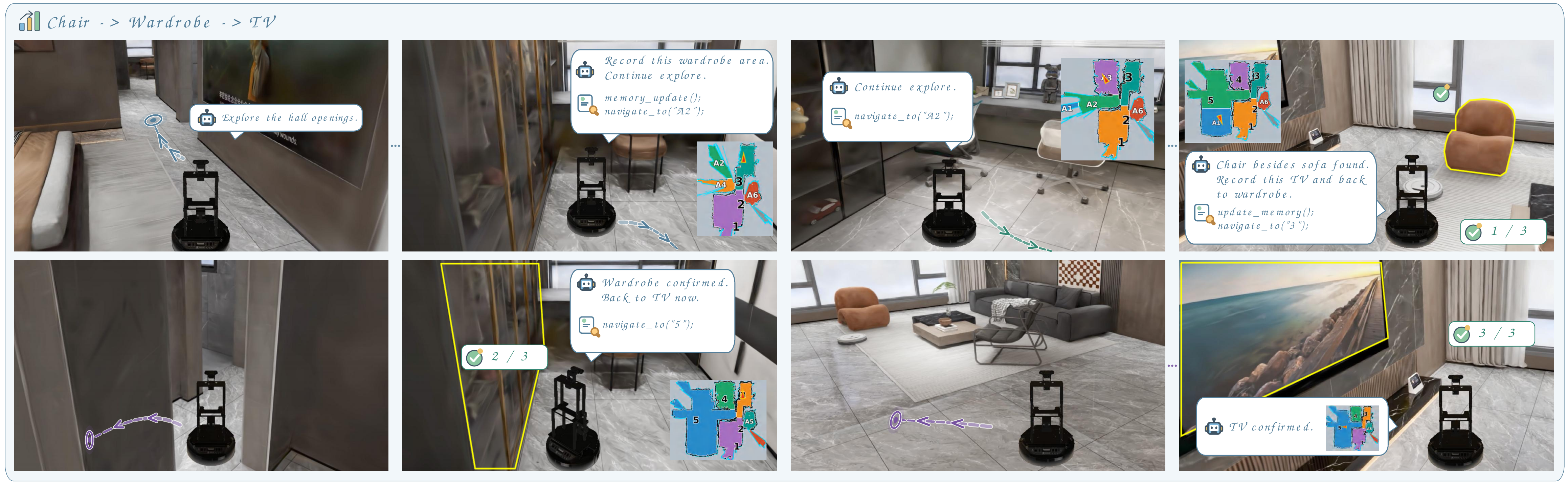}
    \caption{\textbf{Ordered multi-object navigation with spatial memory.} The agent explores unvisited areas with the geometric navigation tool efficiently. Important evidence is recorded by agent and returned to later. }
    \label{fig:multi-object}
\end{figure*}

\paragraph{Descriptions of unvisited areas.}
Intermediate observations encountered during movement may contain useful information and should therefore be retained.
However, for efficiency, agentic navigation systems commonly return only the final observation from a movement-tool call to the reasoning model~\cite{zhang2026supernav,li2026agenticnav}.
We therefore record keyframe images and their poses during execution and asynchronously caption them with Qwen3.5-2B~\cite{qwenteam2026qwen35omnitechnicalreport} (\cref{fig:pipeline}c).
These captions provide textual evidence about areas that the agent has not explicitly visited.

For each unvisited area $a$, we use geometric visibility to select a representative caption.
Let $\mathcal{K}(a)$ denote the set of nearby keyframes, and let $v(k,a)$ count the rays cast from keyframe $k$ that enter area $a$.
We choose
\begin{equation}
    k^{\star}(a) = \operatorname*{arg\,max}_{k\in\mathcal{K}(a)} v(k,a)
    \label{eq:area-keyframe}
\end{equation}
and attach the selected keyframe's caption to the area's map-grounded entry.
These descriptions preserve information gathered during motion as cues for further exploration.

\subsection{Navigation with Memory}
\label{sec:navigation-memory}
The agent uses spatial memory for navigation through three operations. Agent-guided relocalization localizes the robot within the stored map, enabling memory reuse across sessions. Memory lookup retrieves information about remembered places and objects by name or keyword. Geometric navigation then executes an efficient route to the selected destination.

\paragraph{Agent-guided relocalization.}
Reusing spatial memory across sessions requires aligning the robot's new local map with the stored scene map.
At the start of a session, limited geometric observations can make global matching ambiguous when several areas share similar layouts.
To reduce this ambiguity, the agent compares current observations with stored memory descriptions and records to infer its likely area in the prior map (\cref{fig:pipeline}b).
This semantic area prior is then used to constrain geometric matching.

We crop the prior occupancy map to the inferred area and match the current local map against this crop.
The resulting planar rigid transform $\widehat{T}_{\mathrm{prior}\leftarrow\mathrm{new}}$ maps the robot's current pose into the stored map's coordinate frame, making prior memory locations available for navigation.
Restricting the match to the inferred area reduces ambiguity from geometrically similar regions elsewhere in the scene.
The detailed relocalization algorithm and acceptance criteria are provided in the supplementary material (\cref{sec:supp-relocalization}).

\paragraph{Memory lookup.}
The agent retrieves scene information either by node name or through keyword search over stored text.
Direct lookup returns the named node's contents, while keyword lookup returns text-matching entries, as illustrated by the \texttt{grep} example in \cref{fig:pipeline}b.
This allows the agent to inspect relevant descriptions and factual records before selecting a destination, without loading the full memory into its context.

\paragraph{Geometric navigation.}
A geometric navigation tool converts the selected memory or area marker into an executable movement request.
The tool resolves the marker to its associated planar map location and uses A*~\cite{hart1968formal} to plan a path from the robot's current position according to the occupancy map.
The path is converted into primitive actions, which are executed sequentially.
Thus, the agent chooses a destination from scene memory while the geometric tool handles the intervening route, rather than requiring a sequence of locally grounded semantic instructions.
This interface supports navigation to both known locations and unvisited areas marked on the map.
The conversion of an A* path into primitive actions is detailed in the supplementary material (\cref{sec:supp-navigation}).

%% file: sec/experiments.tex
\section{Experiments}
\label{sec:experiments}

\input{tables/hm3d_ovon_results}
\subsection{Experimental Setup}
\label{sec:experimental-setup}
We evaluate \methodname{} for instance navigation on Habitat-GS~\cite{xia2026habitat} and HM3D-OVON~\cite{yokoyama2024hm3dovon}, covering both single- and multi-object tasks.
We evaluate demand-driven navigation on Demand-Bench~\cite{anonymous2026demandagent} and separately test whether the system can reuse its memory across sessions.

Following SuperNav, we use \texttt{GPT-5.6-terra}~\cite{openai2026frontier} as the agent model and the same geometry-based executor.
We report success rate (SR, \%) and success weighted by path length (SPL, $[0,1]$)~\cite{anderson2018evaluation}, where higher values indicate higher completion rates and more efficient trajectories, respectively.
Further evaluation details, additional EQA results, and qualitative demonstrations are provided in the supplementary material (\cref{sec:supp-eval-protocols,sec:supp-addional}).

\begin{figure*}[t!]
    \centering
    \includegraphics[width=\textwidth]{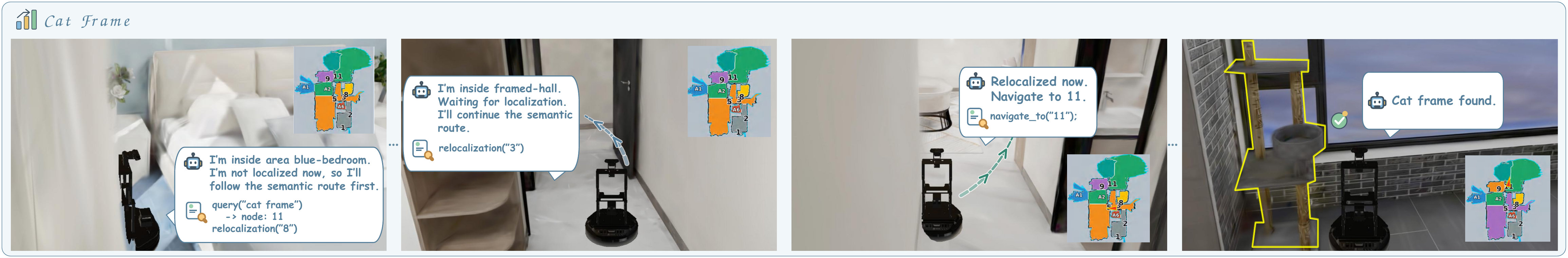}
    \caption{\textbf{Multi-session navigation with memory.} The agent infers its location according to current observations. After relocalization, the geometric navigation tool moves the robot directly to the remembered place.}
    \label{fig:cross-episode}
\end{figure*}

\subsection{Instance Navigation}
\label{sec:instance-navigation}
\input{tables/habitat_gs_results}
We evaluate instance navigation on HM3D-OVON~\cite{yokoyama2024hm3dovon} and Habitat-GS~\cite{xia2026habitat}.

\Methodname{} performs strongly on the 120 episodes in HM3D-OVON val-unseen split (\cref{tab:hm3d-ovon}).
For a fair comparison, we group methods by their evaluation budget.
Among methods evaluated with a 500-step budget, our method achieves the highest reported SR and SPL at the 1\,m threshold. For the 3600\,s agentic comparison, all methods use \texttt{GPT-5.6-terra} as the reasoning model. Our method again achieves the highest reported SR and SPL, outperforming SuperNav, which relies on the agent's interaction history, and HarnessVLN, which maintains event memory and a spatiotemporal graph.

Our method improves more clearly when navigation requires retaining information across multiple targets in Habitat-GS benchmark.
On the 150 single-object tasks in \cref{tab:habitat-gs}, it reaches 79.33\% SR and 0.4246 SPL, exceeding SuperNav on both metrics.
On the 150 multi-object tasks, the improvement is substantially larger: SR increases from 34.00\% to 52.00\%, while SPL rises from 0.1388 to 0.2494. The larger multi-object gain is consistent with the intended role of persistent spatial memory. \cref{fig:multi-object} provides a qualitative example of ordered multi-object navigation.

\input{tables/multi_object_progress}

The target-wise results in \cref{tab:multi-object-progress} show where this gain emerges.
Our method achieves the highest success rates on targets three through five.
At the fifth target, it reaches 60.53\% SR, compared with 39.47\% for SuperNav with the geometry-based executor.
This trend indicates that the advantage accumulates over longer target sequences.

\subsection{Demand-Driven Navigation}
\label{sec:demand-driven}

Unlike instance navigation with explicit object goals, demand-driven navigation requires the agent to infer all objects that satisfy a user need and then navigate to them.
We evaluate \methodname{} on Demand-Bench~\cite{anonymous2026demandagent} in AI2-THOR~\cite{ai2thor}, which contains 70 single-object and 130 multi-object episodes, for 200 episodes in total.
As shown in \cref{tab:demand-driven}, \methodname{} achieves 61.00\% SR and 0.3867 SPL.
The SR gain over SuperNav is modest at 1.50 points, whereas SPL improves by 0.2066, showing a substantially larger gain in navigation efficiency.
\input{tables/demand_driven_results}

\begin{figure*}[t]
    \centering
    \includegraphics[width=\textwidth]{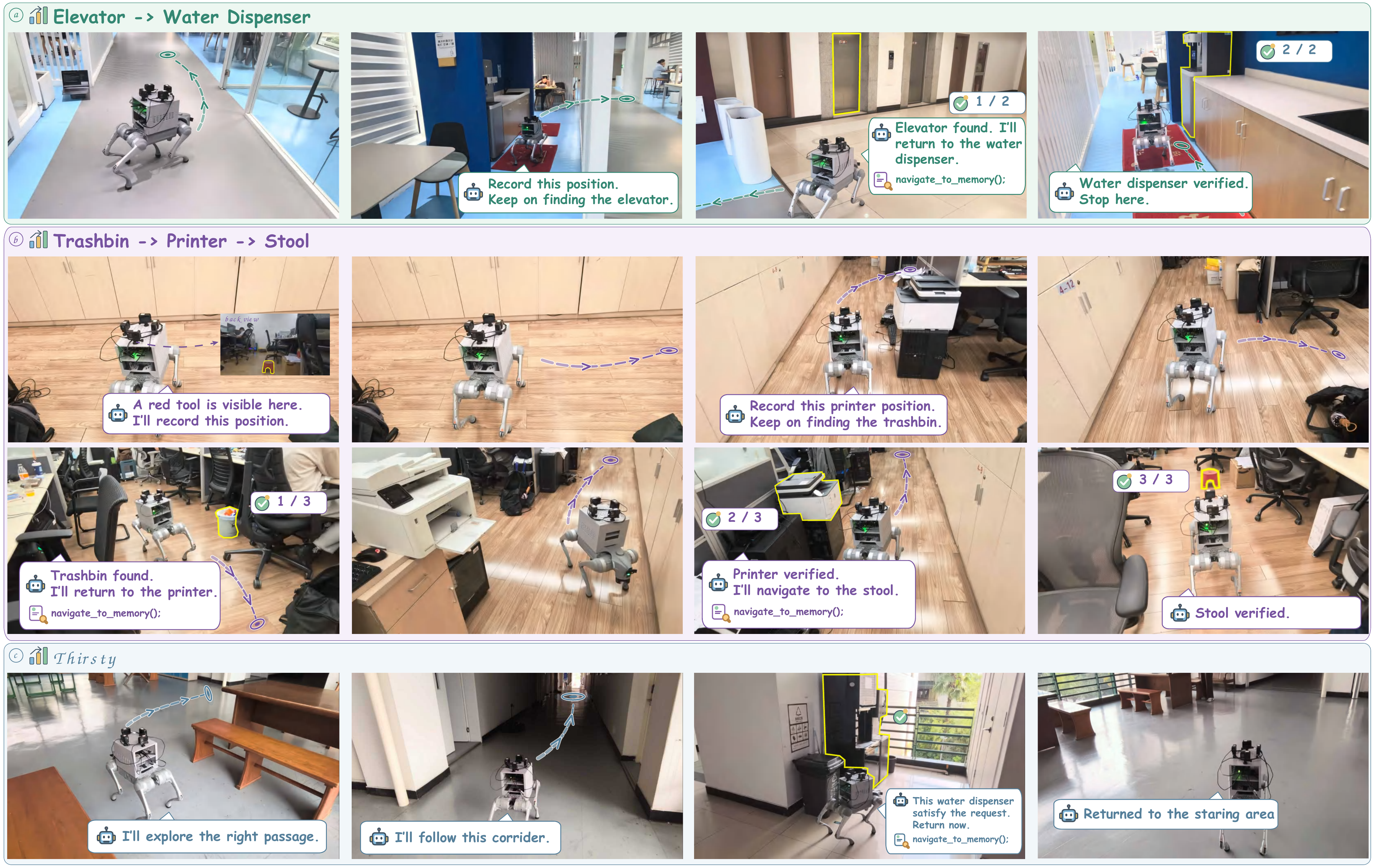}
    \caption{\textbf{Real-world navigation with \methodname{} on a Unitree Go2.}
    (a) The robot records the water dispenser, finds the elevator, and then returns to the remembered dispenser.
    (b) During an ordered trashbin--printer--stool task, it records object locations and later invokes memory-grounded navigation to revisit the printer before proceeding to the stool.
    (c) Given the demand ``thirsty,'' the robot finds a water dispenser and returns to the starting area.}
    \label{fig:real-world-navigation}
\end{figure*}

\subsection{Multi-Session Memory Reuse}
\label{sec:cross-episode}
\input{tables/cross_episode_results}

We next test whether memory acquired in Session 1 improves navigation in Session 2.
In Session 2, the agent inherits the first session's memory and searches for the same objects.
We evaluate both an unchanged spawn position (A) and a different spawn position (B).
In both settings, the agent is not given its initial location and must align its current observations with the stored map before reusing prior memory. We use 150 episodes from the Habitat-GS single-object split as the test set.

The changed-spawn comparison provides the clearest controlled measure of memory reuse (\cref{tab:cross-episode}).
At position B, enabling memory raises SR from 78.00\% to 82.00\% and SPL from 0.4364 to 0.5022.
When both sessions start from position A, the second session likewise improves over the first, from 79.33\% to 82.67\% SR and from 0.4246 to 0.5195 SPL.
The SPL gains in both settings show that the agent reuses prior scene knowledge to navigate more efficiently, even when it begins from a new location. \cref{fig:cross-episode} provides a qualitative example of multi-session navigation.

\subsection{Ablation Study}

\label{sec:cross-episode-ablation}
We conduct three complementary ablation analyses. Using 40 tasks randomly sampled from the Habitat-GS single-object split under the changed-spawn multi-session setting, we first quantify the contributions of the major system components and assess compatibility across reasoning models (\cref{tab:cross-episode-ablation}). We then compare our structured spatial memory with an alternative memory design on all 150 Habitat-GS multi-object tasks.

\paragraph{Main component ablation.}
\input{tables/cross_episode_ablation}
\label{sec:ablation}
For the component ablation, we retain \texttt{GPT-5.6-terra} as the reasoning model.
We first assess whether structured memory nodes must be coupled with a geometric map.
Removing agent-maintaned memory nodes while providing the plain occupancy map reduces SR by 5 points and SPL by 0.0695, providing direct evidence that structued semantic information benefits multi-session navigation.
Removing the annotated occupancy map and the geometric navigation tool for reaching stored locations, while retaining the memory-node tools, reduces SR from 82.50\% to 70.00\% and SPL from 0.5329 to 0.3563, demonstrating the importance of explicit geometry for efficient memory reuse.

We next assess the contribution of agent-guided relocalization.
The \textit{w/o reloc} variant disables both geometric and agent-guided relocalization, leaving the agent with only the static prior map and memory nodes. Its SR and SPL decrease to 72.50\% and 0.3622, respectively, indicating that geometrically localizing the robot within the prior map is necessary for multi-session navigation. The \textit{w/o agent-guided reloc} variant retains global geometric relocalization but removes agent guidance. Its lower SR and SPL show that the agent-provided area prior further improves multi-session navigation.

\paragraph{Reasoning model compatibility.}
To assess compatibility with different reasoning models, we replace \texttt{GPT-5.6-terra} with \texttt{KIMI-k3-256k}~\cite{kimiteam2026kimik3openfrontier} and \texttt{GLM-5.3-flash}~\cite{glm5team2026glm5vibecodingagentic}. \texttt{KIMI-k3-256k} achieves 85.00\% SR and 0.4091 SPL, while \texttt{GLM-5.3-flash} achieves 82.50\% SR and 0.4429 SPL. These results indicate that the navigation system and spatial memory design are not tied to \texttt{GPT-5.6-terra} in this setting.

\input{tables/compare_memory.tex}
\paragraph{Comparison with Another Memory Design.}
To isolate the effect of memory design, we compare our structured spatial memory with that of HarnessVLN~\cite{chen2026harnessvln}, which maintains event memory and a separate spatiotemporal graph. We integrate the HarnessVLN memory design into SuperNav to retain the same camera configuration and geometry-based executor used by \methodname{}. As shown in \cref{tab:compare-memory}, the HarnessVLN memory improves both SR and SPL over the original SuperNav on multi-object navigation, but its gains remain smaller than those achieved by our design.

\subsection{Real-World Demonstrations}
\label{sec:real-world}

We further deploy \methodname{} on a Unitree Go2 in indoor environments. \cref{fig:real-world-navigation} shows three qualitative results.
The first two examples demonstrate multi-object navigation and memory-guided returns to previously observed targets over extended trajectories.
The third shows demand-driven navigation. The robot can interpret ``thirsty,'' locate a water dispenser, and then use memory to return to its starting area.
These qualitative results demonstrate the feasibility of structured memory and geometric execution on a physical platform.

%% file: tables/hm3d_ovon_results.tex
\begin{table}[t]
\caption{\textbf{Results on 120 episodes in HM3D-OVON val-unseen split~\cite{yokoyama2024hm3dovon}.}}
\label{tab:hm3d-ovon}
\centering
\scriptsize
\setlength{\tabcolsep}{2pt}
\begin{tabular*}{\linewidth}{@{\extracolsep{\fill}}lccc@{}}
\toprule
 & & \multicolumn{2}{c}{1\,m} \\
\cmidrule(lr){3-4}
Method & Budget & SR$\uparrow$ & SPL$\uparrow$ \\
\midrule
AstraNav-Memory~\cite{ren2025astranav} & \multirow{3}{*}{500 steps} & 62.50 & 0.3490 \\
OmniNav~\cite{xue2025omninav} & & 59.20 & 0.3320 \\
Ours & & \textbf{78.33} & \textbf{0.4354} \\
\midrule
HarnessVLN~\cite{chen2026harnessvln} & \multirow{3}{*}{$3600\,\mathrm{s}$} & 53.33 & 0.2941 \\
SuperNav~\cite{zhang2026supernav} & & 73.33 & 0.4105 \\
Ours & & \textbf{80.00} & \textbf{0.4317} \\
\bottomrule
\end{tabular*}
\end{table}

%% file: tables/habitat_gs_results.tex
\begin{table}[t]
\caption{\textbf{Single- and multi-object navigation on the Habitat-GS benchmark~\cite{xia2026habitat}.} All methods are evaluated with a 3600\,s time budget. Our method and SuperNav are evaluated using \texttt{GPT-5.6-terra}. Our method achieves the highest SR and SPL.}
\label{tab:habitat-gs}
\centering
\footnotesize
\setlength{\tabcolsep}{3pt}
\begin{tabularx}{\linewidth}{@{}>{\arraybackslash}Xcccc@{}}
\toprule
 & \multicolumn{2}{c}{Single-object (150)} & \multicolumn{2}{c}{Multi-object (150)} \\
\cmidrule(lr){2-3}\cmidrule(lr){4-5}
Method & SR$\uparrow$ & SPL$\uparrow$ & SR$\uparrow$ & SPL$\uparrow$ \\
\midrule
NaVid~\cite{zhang2024navid} & 24.67 & 0.1599 & 2.67  & 0.0216 \\
Uni-NaVid~\cite{zhang2024uninavid} & 34.00 & 0.1833 & 1.33  & 0.0131 \\
StreamVLN~\cite{wei2026streamvln} & 13.33 & 0.1036 & 0.00 & 0.0000 \\
OmniNav~\cite{xue2025omninav} & 27.33 & 0.2133 & 4.00  & 0.0275 \\
SuperNav~\cite{zhang2026supernav} & 78.00 & 0.4127 & 34.00 & 0.1388 \\
Ours & \textbf{79.33} & \textbf{0.4246} & \textbf{52.00} & \textbf{0.2494} \\
\bottomrule
\end{tabularx}
\end{table}

%% file: tables/multi_object_progress.tex
\begin{table}[t]
\caption{\textbf{Progress through ordered multi-object tasks on Habitat-GS~\cite{xia2026habitat}}. $\mathrm{SR}_i$ denotes the success rate at the $i$th target. The number of eligible tasks is shown for each target index.}
\label{tab:multi-object-progress}
\centering
\scriptsize
\setlength{\tabcolsep}{2pt}
\begin{tabular*}{\linewidth}{@{\extracolsep{\fill}}lccccc@{}}
\toprule
\multicolumn{6}{@{}l}{\textit{Multi-object navigation}} \\
Method & $\mathrm{SR}_{1}\uparrow$ & $\mathrm{SR}_{2}\uparrow$ & $\mathrm{SR}_{3}\uparrow$ & $\mathrm{SR}_{4}\uparrow$ & $\mathrm{SR}_{5}\uparrow$ \\
Eligible tasks & 150 & 150 & 112 & 75 & 38 \\
\midrule
NaVid~\cite{zhang2024navid} & 40.00 & 6.67 & 2.68 & 1.33 & 0.00 \\
Uni-NaVid~\cite{zhang2024uninavid} & 36.67 & 4.00 & 1.79 & 1.33 & 0.00 \\
StreamVLN~\cite{wei2026streamvln} & 22.00 & 3.33 & 0.89 & 0.00 & 0.00 \\
OmniNav~\cite{xue2025omninav} (Action Former) & 44.00 & 11.33 & 1.79 & 1.33 & 0.00 \\
SuperNav~\cite{zhang2026supernav} + Learned Executor & \textbf{86.00} & \textbf{63.33} & 52.68 & 42.67 & 34.21 \\
SuperNav~\cite{zhang2026supernav} + Geo-based Executor & 85.33 & 62.00 & 48.21 & 38.67 & 39.47 \\
Ours & 82.67 & \textbf{63.33} & \textbf{59.82} & \textbf{56.00} & \textbf{60.53} \\
\bottomrule
\end{tabular*}
\end{table}

%% file: tables/demand_driven_results.tex
\begin{table}[t]
\caption{\textbf{Demand-driven navigation on Demand-Bench~\cite{anonymous2026demandagent}} Total 70 single-object and 130 multi-object episodes. All methods are evaluated with a 3600\,s time budget. Our method and SuperNav are evaluated using \texttt{GPT-5.6-terra}. Our method achieves the highest SR and SPL.}
\label{tab:demand-driven}
\centering
\footnotesize
\setlength{\tabcolsep}{4pt}
\begin{tabular*}{0.85\linewidth}{@{\extracolsep{\fill}}lcc@{}}
\toprule
Method & SR$\uparrow$ & SPL$\uparrow$ \\
\midrule
NaVid~\cite{zhang2024navid} & 17.00 & 0.0920 \\
Uni-NaVid~\cite{zhang2024uninavid} & 25.00 & 0.0943 \\
StreamVLN~\cite{wei2026streamvln} & 25.50 & 0.0775 \\
OmniNav~\cite{xue2025omninav} & 37.50 & 0.1213 \\
SuperNav~\cite{zhang2026supernav} & 59.50 & 0.1801 \\
\midrule
Ours & \textbf{61.00} & \textbf{0.3867} \\
\bottomrule
\end{tabular*}
\end{table}

%% file: tables/cross_episode_results.tex
\begin{table}[t]
\caption{\textbf{Multi-session performance under unchanged (A) and changed (B) spawn positions.} Session 2 reuses memory from Session 1 if marked ``with memory.'' Memory improves multi-session performance whether the spawn position is unchanged or changed between sessions.}
\label{tab:cross-episode}
\centering
\footnotesize
\setlength{\tabcolsep}{4pt}
\begin{tabular*}{0.95\linewidth}{@{\extracolsep{\fill}}lccc@{}}
\toprule
Session & Spawn Position & SR$\uparrow$ & SPL$\uparrow$ \\
\midrule
Session 1 & A & 79.33 & 0.4246 \\
Session 2 (with memory) & A & \textbf{82.67} & \textbf{0.5195} \\
\midrule
Session 2 (no memory) & B & 78.00 & 0.4364 \\
Session 2 (with memory) & B & \textbf{82.00} & \textbf{0.5022} \\
\bottomrule
\end{tabular*}
\end{table}

%% file: tables/cross_episode_ablation.tex
\begin{table}[t]
\caption{\textbf{Multi-session ablation of memory node, occupancy mapping, relocalization, and reasoning models.}}
\label{tab:cross-episode-ablation}
\centering
\footnotesize
\setlength{\tabcolsep}{4pt}
\begin{tabular*}{0.6\linewidth}{@{\extracolsep{\fill}}lcc@{}}
\toprule
Variant & SR$\uparrow$ & SPL$\uparrow$ \\
\midrule
w/o memory nodes & 77.50 & 0.4634 \\
w/o occupancy map & 70.00 & 0.3563 \\
w/o reloc  & 72.50 & 0.3622 \\
w/o agent-guided reloc & 77.50 & 0.4381 \\
Ours (GPT-5.6-terra \cite{openai2026frontier}) & \textbf{82.50} & \textbf{0.5329} \\
\midrule
Ours (KIMI-k3-256k \cite{kimiteam2026kimik3openfrontier}) & 85.00 & 0.4091 \\
Ours (GLM-5.3-flash \cite{glm5team2026glm5vibecodingagentic}) & 82.50 & 0.4429 \\
\bottomrule
\end{tabular*}
\end{table}

%% file: tables/compare_memory.tex
\begin{table}[t]
\caption{\textbf{Comparison of memory designs on Habitat-GS multi-object navigation.}}
\label{tab:compare-memory}
\centering
\footnotesize
\setlength{\tabcolsep}{4pt}
\begin{tabular*}{0.85\linewidth}{@{\extracolsep{\fill}}lcc@{}}
\toprule
Method & SR$\uparrow$ & SPL$\uparrow$ \\
\midrule
SuperNav~\cite{zhang2026supernav} & 34.00 & 0.1388 \\
SuperNav + HarnessVLN ~\cite{chen2026harnessvln} & 35.33 & 0.1797 \\
SuperNav + Ours & \textbf{52.00} & \textbf{0.2494} \\
\bottomrule
\end{tabular*}
\end{table}

%% file: sec/conclusion.tex
\section{Conclusion}
We propose \methodname, an agentic navigation system equipped with structured spatial memory for efficient navigation across sessions and successive object goals.
The annotated occupancy map provides a unified interface through which the agent accesses spatial memory, while agent-maintained memory nodes selectively retain scene knowledge.
Experiments on instance-navigation and demand-driven benchmarks, as well as in multi-session settings, show that our method achieves higher success rates and greater path efficiency.
\paragraph{Limitations.} The occupancy map and corresponding geometric navigation tool depend on accurate LiDAR or depth observations and odometry estimates. Sensor errors or odometry drift may distort the reconstructed geometry. Improving the long-term stability of this geometry is an important direction for future work.

%% file: sec/supplementary.tex
\clearpage
\setcounter{page}{1}
\maketitlesupplementary

\section{Evaluation Protocols}
\label{sec:supp-eval-protocols}
\subsection{Shared Evaluation Controls}
\label{sec:supp-eval-controls}

Unless stated otherwise, all navigation experiments use \texttt{GPT-5.6-terra} with high reasoning effort. Alternative-model ablations retain high reasoning effort and change only the policy model. Each episode is run once in an isolated session. The agent receives only the public task instruction, RGB observations, and tools enabled for that condition. Target coordinates, evaluator geometry, approved viewpoints, reference paths, and success feedback remain hidden from the agent.

An episode is complete only when the agent issues an explicit \textsc{Stop} action. Failed episodes because of infrastructure failures, such as simulator or upstream model provider errors, are retried.

\subsection{Dataset-Specific Evaluation}
\label{sec:supp-dataset-evaluation}

\paragraph{InteriorGS single-object navigation.}
An episode succeeds when the agent issues \textsc{Stop} and its final saved position is at a navmesh-geodesic distance strictly below $1.0\,\mathrm{m}$ from at least one approved viewpoint for the target. Only the final position determines success. Encountering the target earlier does not count if the agent subsequently moves away. We report success rate (SR) and success weighted by path length (SPL), where the per-episode SPL is
\begin{equation}
\mathrm{SPL}_i=S_i\frac{L_i}{\max(L_i,P_i)}.
\end{equation}
Here, $S_i$ is the binary success indicator, $L_i$ is the frozen shortest-path reference distance, and $P_i$ is the accumulated length of the saved trajectory through \textsc{Stop}. We compute path length directly from consecutive saved poses without trajectory interpolation. All methods are evaluated with a 3600\,s time budget.

\paragraph{InteriorGS ordered multi-object navigation.}
Targets must be completed in the specified order. The evaluator scans the saved trajectory chronologically and completes the active target at the first recorded position whose navmesh-geodesic distance to an approved target viewpoint is strictly below $1.0\,\mathrm{m}$. A single position may complete multiple consecutive targets when it lies within all corresponding acceptance regions. Visiting a later target before it becomes active does not complete that target. Full-task success requires completing every target and then issuing \textsc{Stop}. All methods are evaluated with a 3600\,s time budget.

Besides full-task SR and sequential SPL, we report the ordered completion fraction and prefix metrics $\mathrm{SR}_k$ and $\mathrm{SPL}_k$. The $k$-th prefix cohort includes every episode containing at least $k$ targets, rather than only episodes containing exactly $k$ targets.

\paragraph{Demand-bench.}
Demand-bench evaluates roles in the serialized order of the released \texttt{goal\_graph}. At every saved pose, including the initial pose, the evaluator considers only the active role and selects its nearest unused admissible object. Exact distance ties are resolved by object identifier. Distance is measured in the horizontal plane from the agent's $(x,z)$ position to the nearest point on the object's runtime AI2-THOR axis-aligned bounding-box footprint. The active role is completed when this distance is at most $2.0\,\mathrm{m}$. An object cannot satisfy more than one role, although one pose may complete several consecutive roles using different objects.

Acceptance depends only on this geometric rule. Full-task success requires completing every role and issuing \textsc{Stop}. Path length is accumulated in the horizontal plane, and SPL uses the frozen globally optimal route for the complete task.
All methods are evaluated with a 3600\,s time budget.

\paragraph{HM3D-OVON.}
Success is determined by the minimum navmesh-geodesic distance from the final saved position to any approved viewpoint of any instance belonging to the target category.
We report SR and SPL at the strict navmesh-geodesic threshold $d<1.0\,\mathrm{m}$, requiring an explicit \textsc{Stop}.
For SPL, the reference length is the start-to-goal navmesh-geodesic distance stored in the original HM3D-OVON episode, whereas the executed path length is accumulated over the complete saved trajectory from the initial position to the pose at which \textsc{Stop} is issued.

Three episodes contain targets that are unreachable on the navigation mesh.
We count these episodes as successful and assign each an $\mathrm{SPL}$ of 1.

\subsection{Multi-session Evaluation Protocols}
We evaluate multi-session performance on all 150 episodes in the Habitat-GS single-object split. Each episode comprises two sequential sessions. Unless an experimental condition explicitly disables memory, Session 2 receives the spatial memory frozen at the end of Session 1, including the occupancy map and associated memory nodes. We consider two Session 2 spawn settings: the unchanged-spawn setting reuses the Session 1 spawn position, whereas the changed-spawn setting uses a different position. For each episode, we sample this alternative position once before running any experiments. The sampled position is guaranteed to be navigable, not initially reveal the target, and have a ground-truth path length to the target of at least half that from the Session 1 spawn. Once selected, the position remains fixed across all experimental conditions.

\subsection{Ablation Protocols}
\label{sec:supp-ablation-protocols}

We conduct the ablation study on a randomly sampled 40-task subset of the Habitat-GS single-object split. All experiments use the changed-spawn multi-session setting: in Session 2, the robot starts from a location different from its Session 1 spawn and receives the prior spatial memory, except when an ablation removes specific memory components.
We evaluate the following configurations:
\begin{description}[leftmargin=*,style=nextline]
    \item[Baseline.] Loads the paired memory nodes, visual and planning SLAM states, caption evidence, and relocalization machinery. This is the reference condition.
    \item[No occupancy map.] The annotated occupancy map is hidden from the agent, and geometric navigation to known locations is disabled. Visual-point navigation and memory nodes remain available.
    \item[No memory node.] Memory-node lookup and update are disabled, as is geometric navigation to memory-node locations. The paired SLAM maps and caption evidence remain available, isolating geometric mapping from structured memory.
    \item[No relocalization.] Static memory nodes and the occupancy map from the first session are retained, but both agent-triggered and geometric relocalization are disabled. No initial pose is inserted into the inherited map.
    \item[No agent-guided relocalization.] The agent-guided relocalization pathway is disabled. Background geometric-based relocalization remain enabled. This condition isolates the contribution of the semantic prior from automatic geometric relocalization.
\end{description}

\paragraph{Alternative policy models.}
The model ablation changes only the policy models. The completed comparison cells use \texttt{KIMI-k3-256k} and \texttt{GLM-5.3-flash}.

\subsection{HarnessVLN Memory Integration}
To compare memory designs without confounding them with different agent harnesses, we integrate HarnessVLN's memory~\cite{chen2026harnessvln} into SuperNav and evaluate both memory systems under the same SuperNav runtime. This comparison uses all 150 Habitat-GS multi-object episodes with identical initial poses and ordered target sequences. The two conditions also share the reasoning-model settings, four-camera RGB observations, action and time budgets, and objective evaluator. They differ only in the memory representation, update protocol, and retrieved context exposed to the agent.

\paragraph{Core memory components.}
We preserve HarnessVLN's two core memory components. Its event memory maintains a structured TODO list with stable identifiers, four states (\texttt{PENDING}, \texttt{ACTIVE}, \texttt{COMPLETED}, and \texttt{BLOCKED}), at most one active item, incremental updates, and evidence references. Its episode-local spatiotemporal graph retains place, object, failure, and edge records. Places within $0.75\,\mathrm{m}$ are merged, as are objects within $1.0\,\mathrm{m}$. Objects within $0.35\,\mathrm{m}$ may also be merged despite having different labels, with those labels retained as aliases. The implementation preserves the \texttt{NAVIGABLE\_TO}, \texttt{CONTAINS}, \texttt{OBSERVED\_FROM}, and \texttt{FAILED\_FOR\_SUBGOAL} relations and adds navigability edges only after an actual traversal. It also records place summaries and the subgoal-specific \texttt{NO\_PROGRESS} and \texttt{STOP\_REJECTED} failures.

\paragraph{Memory retrieval.}
Memory retrieval runs automatically before every high-level decision and uses HarnessVLN's retrieval scoring function. It selects the three highest-scoring content-bearing places as seeds, expands one hop from these seeds and the current place, and returns at most five places while excluding the current place itself. The agent then commits its summary of the current place, TODO updates, and intended action before execution.

\paragraph{Navigation capability.}
To make recalled information actionable while holding navigation capability fixed across experimental conditions, the adapter associates HarnessVLN's graph places with poses from our Cartographer state. For a visible target, the agent invokes SuperNav's point-only \texttt{visual\_point\_navigate} interface. For a recalled place, it invokes our \texttt{navigate\_slam\_occupancy} interface, which uses the shared occupancy-map planner. Backtracking retains HarnessVLN's actionability rule. The agent may select only a waypoint returned by the current retrieval that lies within $6\,\mathrm{m}$. Thus, all experimental conditions share the same Cartographer state, local point-navigation interface, occupancy-based navigation, and low-level execution. The comparison primarily isolates the memory representation and retrieval protocol.

\begin{figure*}[t]
    \centering
    \includegraphics[width=\textwidth]{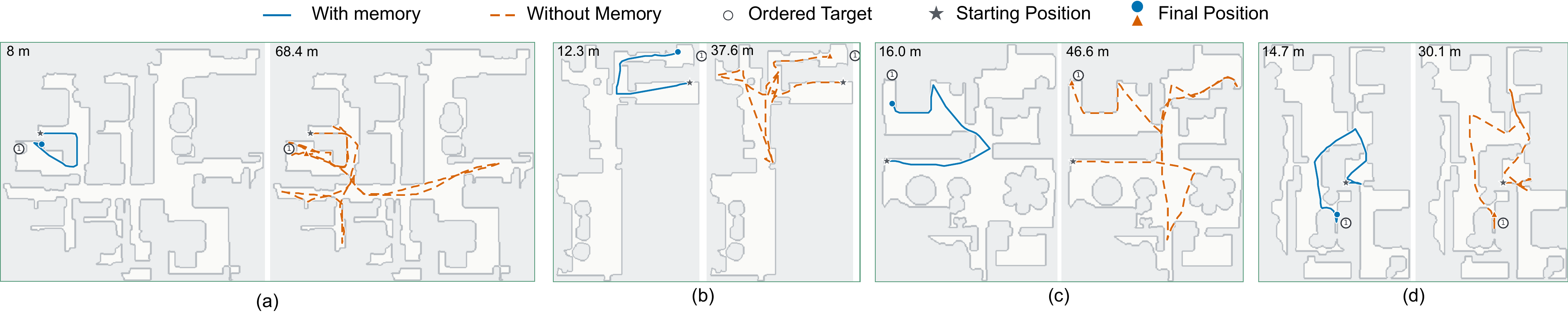}
    \caption{\textbf{Navigation trajectories in the changed-spawn multi-session setting.} By reusing memory from Session 1, the agent follows a shorter, more direct route to the target in Session 2.}
    \label{fig:memory_compare_ablation}
\end{figure*}

\paragraph{Tradeoffs.}
The integration necessarily omits components that are incompatible with the shared harness or are unrelated to the memory design. Official HarnessVLN explicitly replays a thinned history of up to 50 images at each planning request, whereas the adapted condition guarantees only SuperNav's latest set of four camera views. Older images may remain in the persistent chat context but are not explicitly replayed as memory input. We also omit HarnessVLN's semantic map, whose route-planning role is instead fulfilled by our shared occupancy map, as described above. HarnessVLN's remaining navigation and stopping components, including its separate VLM grounding call and FMM/NavDP controller, are replaced by the corresponding SuperNav implementations. These adaptations preserve its structured event memory, spatiotemporal graph, and retrieval behavior while avoiding differences in perception, control, and evaluation infrastructure.

\section{Additional Results}
\label{sec:supp-addional}
\subsection{Embodied Question Answering}
\label{sec:supp-eqa}

Embodied question answering (EQA) requires an agent to explore an environment and gather visual evidence before answering a question. Because the required evidence may occur in different areas and at different points along the trajectory, EQA also tests whether the agent can retain and combine observations collected during exploration. We evaluate this capability on 150 EQA tasks in Habitat-GS~\cite{zhang2026supernav}. Their answer- and question-type distributions are summarized in \cref{tab:eqa-breakdown}.

\paragraph{Evaluation protocol.}
All methods except Uni-NaVid use \texttt{GPT-6.1-sol} with medium reasoning effort. Uni-NaVid uses its native 7B policy model. Question is provided at the start of each task. The first valid \texttt{hab\_eqa\_stop} action terminates an episode and supplies the predicted answer. Single-choice questions require one correct label, whereas multi-choice questions require the exact set of unique labels. Integer answers exclude Boolean values, and yes/no questions accept Boolean values or normalized yes/no strings. Unordered sets require exact membership, and ordered lists additionally require the correct order. Normalization is limited to Unicode NFKC normalization, case folding, and whitespace normalization. Synonyms and partial matches receive no credit. A missing or invalid answer, failure to stop, or navigation-budget exhaustion is scored as incorrect, while failed episodes caused by simulator or upstream model provider are retried.

\input{tables/eqa_type}

\begin{figure*}[t]
    \centering
    \includegraphics[width=\textwidth]{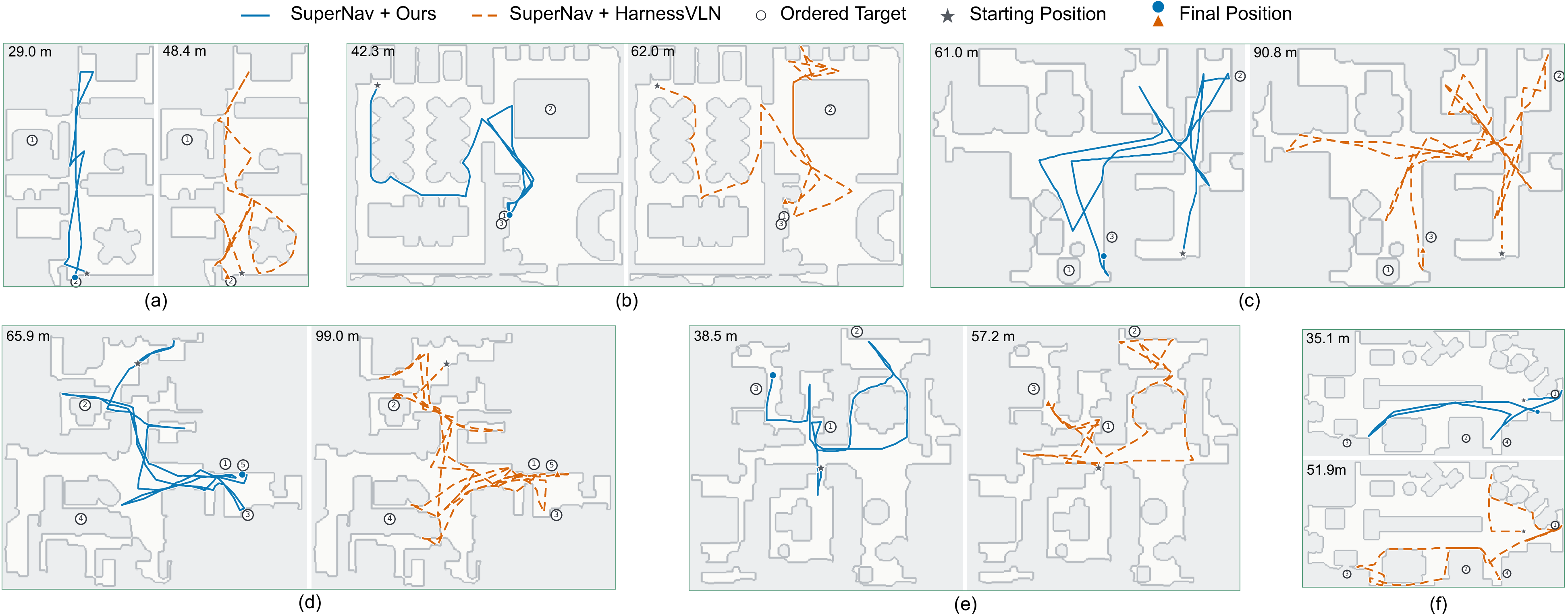}
    \caption{\textbf{Navigation trajectories using our memory and HarnessVLN's memory design.} In these multi-object tasks, our memory enables shorter trajectories with fewer redundant motions.}
    \label{fig:memory_compare_harnessvln}
\end{figure*}
\begin{figure*}[t]
    \centering
    \includegraphics[width=\textwidth]{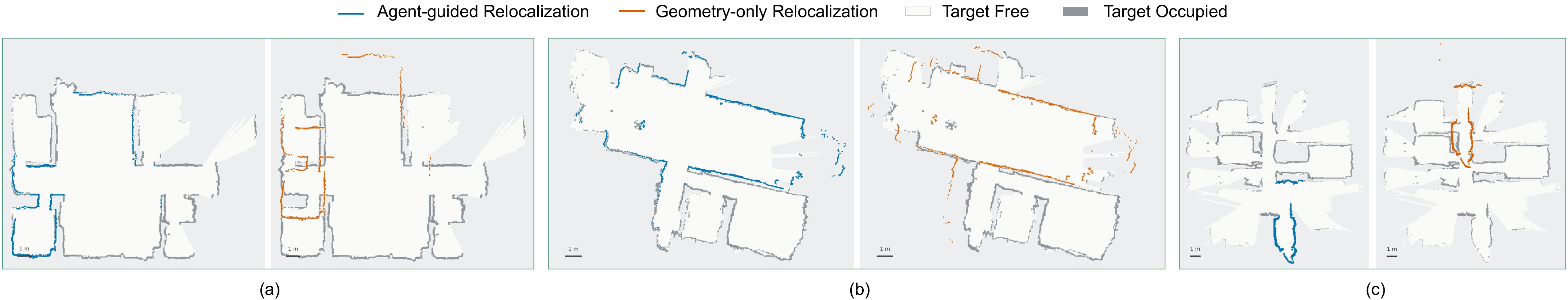}
    \caption{\textbf{Agent-guided versus geometry-only relocalization.} Agent guidance resolves ambiguities caused by similar wall structures and limited geometric observations, producing more accurate alignment with the stored map.}
    \label{fig:reloc_compare}
\end{figure*}

\paragraph{Overall comparison.}
As shown in \cref{tab:eqa-overall}, our method answers 131 of 150 questions correctly, achieving 87.33\% accuracy. This is 6.00 percentage points above SuperNav and the highest accuracy among the evaluated methods. Since SuperNav uses the same reasoning model, this comparison indicates that the added structured spatial memory and its associated navigation mechanisms improve the agent's ability to collect and retain task-relevant evidence.

\subsection{Qualitative Results on Path Efficiency}
We visualize navigation trajectories to illustrate how spatial memory improves path efficiency.
\paragraph{Comparison in multi-session experiment.}
We compare Session 2 trajectories under the changed-spawn setting described in \cref{sec:cross-episode}. In \cref{fig:memory_compare_ablation}, blue trajectories use memory inherited from Session 1, whereas orange trajectories start without inherited memory. Across these examples, the memory-enabled agent reaches the target along shorter, more direct routes and avoids unnecessary exploratory detours. These results illustrate that memory acquired in an earlier session can be reused to improve subsequent navigation efficiency.
\paragraph{Comparison with another memory method.}
We further compare our memory with HarnessVLN's memory design on the multi-object navigation tasks described in \cref{sec:ablation}. In \cref{fig:memory_compare_harnessvln}, blue trajectories show SuperNav with our memory, whereas orange trajectories show SuperNav with HarnessVLN's memory. HarnessVLN sometimes revisits areas that contain no active target. For example, it repeatedly explores the upper-right room in panel (c). When an agent encounters a later target early in an episode, our memory preserves its location and supports a more direct return when that target becomes active, whereas the HarnessVLN variant may produce redundant motion.
\subsection{Qualitative Results on Agent-guided Relocalization}
We qualitatively compare agent-guided and geometry-only relocalization using cases from the multi-session experiment in \cref{sec:cross-episode}. In Session 2, the agent must estimate its pose within the stored map before it can geometrically reuse memory from Session 1. In \cref{fig:reloc_compare}, the gray background represents the target occupancy map stored during Session 1. The blue and orange points show the current-session occupancy map transformed using agent-guided and geometry-only relocalization, respectively. Panels (a) and (b) show that geometry-only matching can select an incorrect alignment when the environment contains similar wall structures. Panel (c) further illustrates its ambiguity when the current occupancy map contains limited geometric structure. In these cases, the agent-provided spatial prior reduces ambiguity and yields a more accurate alignment with the stored map.

\section{Implementation Details}
\label{sec:supp-implementation}

This section describes the construction of the annotated occupancy map, the persistent memory-node interface, and the geometric navigation executor.

\subsection{Occupancy-Map Construction}
\label{sec:supp-occupancy}

\paragraph{Occupancy map in simulation}
In simulation, we construct two geometrically aligned occupancy maps from the same timestamped simulator odometry and depth observations: a \emph{center-band structural map} and a \emph{projected-depth planning map}.
The two maps serve different accuracy requirements. The structural map favors clean room boundaries for area segmentation and relocalization, whereas the planning map favors collision-relevant obstacle coverage for path planning and execution.

For the structural map, we sample a narrow horizontal band around the image center and convert its depths into angularly spaced planar ranges. This camera-height slice captures walls and other structures intersecting the sampling plane and generally produces clean free-space boundaries. It may, however, miss low furniture and obstacles visible only in the lower image. The planning map therefore back-projects pixels from the lower part of the depth image into 3D, retains points between $0.15$ and $1.25\,\mathrm{m}$ above the floor, and projects them onto the ground plane. The nearest point in each angular bin becomes the simulated LiDAR return. Keeping the two streams separate avoids contaminating the structural map with dense object boundaries while retaining the obstacle coverage needed for safe planning.

The simulator pose is transformed into Cartographer's planar tracking frame and rebased to the first observation. A patched Cartographer local trajectory builder uses timestamp-matched simulator odometry as the authoritative trajectory pose: it bypasses local scan-pose estimation and inserts each range observation at the supplied odometry pose. Because the two scan streams share timestamps and odometry, their grids remain aligned. Cartographer integrates the scans into submaps and renders both grids at a resolution of $\rho=0.05\,\mathrm{m}$ per cell.

\paragraph{Occupancy map in real-world deployment.}
The real-world occupancy map is a world-aligned, terrain-aware two-dimensional costmap constructed from the Unitree Go2's onboard L1 LiDAR and odometry. It provides the common geometric representation used for map visualization, area segmentation, and navigation.

The robot interface receives LiDAR point arrays and robot odometry over WebRTC. Each LiDAR observation is a 3D point cloud expressed in the Unitree world frame, and odometry provides the corresponding world-to-\texttt{base\_link} pose for tracking and displaying the robot. The mapper inserts these already registered world-frame observations directly, without additional local scan matching or pose optimization. Mapper follows the implementation in dimos \cite{dimos2026}.

The world-aligned costmap supplies both the displayed geometry and the planner's input. Before planning, lethal cells are inflated by $0.15\,\mathrm{m}$, half of the Go2's nominal $0.30\,\mathrm{m}$ width, and an obstacle-distance cost field encourages paths with additional clearance.

\paragraph{Area segmentation.}
The structural occupancy grid is partitioned into human-interpretable areas in four stages.
\begin{enumerate}[leftmargin=*,label=\arabic*.]
    \item \textbf{Cell classification and cleanup.} We classify grayscale intensities below 80 as occupied, intensities of at least 245 as free, and all remaining values as unknown. We then extract eight-connected free-space components and discard components smaller than $A_{\min}=80$ pixels.
    \item \textbf{Conditional subdivision.} A component smaller than $\max(4A_{\min},400)$ pixels is retained as one area. Larger components are candidates for subdivision because they may join several rooms through narrow openings. For each candidate, we compute the Euclidean distance $D$ from every free cell to the component boundary. If the maximum clearance is below three pixels, we retain the component intact. This avoids fragmenting thin corridors that contain no reliable room interior.
    \item \textbf{Watershed growth and stability merging.} We smooth $D$ with a Gaussian kernel of $\sigma=1$ and detect local-maximum pixels using an adaptive odd window between 15 and 65 pixels. Only maximum pixels with at least three pixels of clearance are retained. Because a flat peak may contain several adjacent maximum pixels, we group the retained pixels into eight-connected components. Each component becomes one watershed seed and receives a unique integer basin label. We then process the remaining free cells in descending order of smoothed clearance and propagate these basin labels through four-neighbor connections. Thus, every assigned cell carries the identifier of the seed from which its basin grew, and neighboring basins meet near doorways or other narrow boundaries. After this initial partition, we evaluate each basin using its area, maximum clearance, thick-interior support, occupied/unknown boundary ratios, compactness, and contact with neighboring basins. A basin that fails the stability test is absorbed into the adjacent basin with which it shares the longest boundary. Final basins smaller than $A_{\min}$ are discarded.
    \item \textbf{Area metadata.} For every retained area, we store its pixel count, bounding box, centroid, number of free--unknown frontier cells, and boundary size.
\end{enumerate}
The result is an area-label grid together with area records. The conditional subdivision preserves simple or narrow spaces, while the stability merge prevents the raw watershed from producing many small regions that would be difficult for the agent to interpret.

\paragraph{Captioning and geometric evidence association.}
Successful robot movements produce a caption keyframe every two translation steps. Each keyframe stores front, right, back, and left RGB views, which are submitted to an asynchronous caption inference queue.

We use Qwen3.5-2B~\cite{qwenteam2026qwen35omnitechnicalreport} as the captioning backend. Before inference, images are resized to at most 768 pixels on the longest side. Captions are generated deterministically with at most 48 new tokens. The prompt requests one short factual description of visible objects, surfaces, openings, and spatial layout, while explicitly prohibiting guesses about room category, hidden objects, navigation goals, or actions. This restriction favors concise visual evidence over unsupported semantic inference.

We associate captions with segmented areas in two stages. If at least one keyframe pose lies inside an area, we select the keyframe closest to the area centroid and attach all four directional captions. When multiple candidates share the same position, preference is given to current-session and more recent evidence. If no keyframe lies inside the area, we use ray casting to select the view that best depicts the area. For every directional view, nine rays are distributed uniformly across its $90^\circ$ horizontal field of view. Each ray advances until it leaves the grid, encounters an occupied cell, or enters the candidate area. We select the view with the most rays entering the area, breaking ties by distance to the area center, and recency, and attach only that view's caption. These captions remain transient visual evidence and are not automatically promoted to persistent semantic memory.

\subsection{Persistent Memory Nodes}
\label{sec:supp-memory-nodes}

\paragraph{Data structure.}
Persistent memory is stored as a JSON document containing a set of node records. Each agent-authored node contains a stable \texttt{id}, a \texttt{kind} (\texttt{area} or \texttt{door}), a short \texttt{name}, a semantic \texttt{description}, a list of textual \texttt{records}, and \texttt{pano\_images} mapping view directions to panorama references. The system additionally maintains a monotonically increasing \texttt{order\_id} and a \texttt{slam\_pose} containing the Cartographer sequence, planar position, and yaw at which the node was recorded. Updating panoramic evidence also refreshes the SLAM pose when a current pose is available.

\paragraph{Creation and update.}
Creation requires a complete node and rejects duplicate identifiers. A partial update may modify only \texttt{name}, \texttt{description}, \texttt{records}, and \texttt{pano\_images}. After any mutation, the system republishes the node manifest used to annotate the occupancy map.

\paragraph{Lookup.}
The interface provides three complementary access patterns. Listing returns compact summaries of all nodes containing the name, and description. Text search performs case-insensitive substring matching over names, descriptions, and textual records. Inspection returns a complete node information and can omit panorama references when images are not required. A combined context query can execute multiple searches and inspections against the memory.

\paragraph{Atomic batch mutation.}
Batch mutation accepts an ordered, nonempty sequence of \texttt{create}, \texttt{create}, and \texttt{update} operations. The system validates and applies all operations to an in-memory copy before performing one persistent write. Its response reports the index, status, and node identifier of every operation, followed by image-free snapshots of the affected nodes. This batch mutation saves agent steps during navigation.

\subsection{Agent-Guided Relocalization}
\label{sec:supp-relocalization}

Agent-guided relocalization aligns geometry observed in the current session (the \emph{source}) with the stored occupancy map (the \emph{target}). At the beginning of a session, the current trajectory may contain too little geometry for reliable global matching. The agent therefore supplies a semantic spatial prior by recognizing a stored memory node, after which geometric registration estimates and verifies the precise planar transform. The procedure consists of region construction, source-map preparation, candidate search and refinement, conservative acceptance, and frame handoff.

\paragraph{Agent-guided target region.}
When the agent recognizes a previously stored node, it reports that node as the relocalization prior. We locate the node's stored SLAM pose in the segmented target occupancy map. The free-space area containing the pose becomes the primary target region. If the pose is outside all segmented areas, we use the nearest area. We then include any area whose bounding box lies within $0.5\,\mathrm{m}$ of the reported node and take the union of the selected area bounds. Padding this union by $0.5\,\mathrm{m}$ produces the final metric crop. If no segmentation is available, the fallback region is an $8\times8\,\mathrm{m}$ square centered on the node pose.

The crop restricts only the occupied and free cells of the stored target map. We retain the complete recent source trajectory so that all currently observed geometry can support alignment. For each candidate transform, we transform the current robot pose---the endpoint of the source trajectory---into the target-map frame and require it to lie inside the agent-reported region.

\paragraph{Current-session source geometry.}
The source representation uses the most recent $8\,\mathrm{m}$ of scan-matched trajectory, measured along the traveled path. We insert its scans into a temporary Cartographer-style probability grid at $0.05\,\mathrm{m}$ resolution, using both ray hits and misses. Cells with occupancy probability at least 0.65 form the occupied source set, whereas cells with probability at most 0.46 form the confidently free set.

\paragraph{Candidate generation and refinement.}
We generate initial source-to-target transforms using multi-scale FPFH/RANSAC registration. Each proposal is expanded by $90^\circ$ quadrant rotations and local yaw perturbations, then pre-ranked using both point-cloud fitness and consistency between the transformed source occupancy and the target grid. Correlative scan matching searches around the strongest proposals. A coarse stage covers translations of $\pm1.5\,\mathrm{m}$ and rotations of $\pm15^\circ$. A fine stage then searches $\pm0.10\,\mathrm{m}$ and $\pm2^\circ$. Surviving candidates are refined with planar point-to-point ICP and projected to rigid $\mathrm{SE}(2)$ transforms. This sequence combines broad hypothesis generation with progressively finer grid and point-cloud alignment.

\paragraph{Acceptance gates.}
A candidate transform is accepted only when it satisfies every gate below. The gates test whether enough geometry is available, whether occupied and free space agree, whether matches cover the trajectory rather than one small patch, whether local refinement is reliable, and whether the result remains consistent with the agent's prior.

We compute occupancy consistency only where both maps provide a confident state. Let $O_s$ and $F_s$ denote the occupied and free source cells, respectively. For a candidate transform $T$, a source cell is \emph{comparable} if its transformed center falls inside the target crop and the corresponding target cell is confidently occupied or free. Target-unknown cells and cells outside the crop are excluded. Let $O_s^c(T)$ and $F_s^c(T)$ be the resulting comparable subsets, and let $\ell_t(\cdot)\in\{\mathrm{occ},\mathrm{free}\}$ denote the target-cell state. We define the two conflict ratios as
\begin{align}
r_{o\rightarrow f}(T)=
\frac{\left|\{p\in O_s^c(T):\ell_t(Tp)=\mathrm{free}\}\right|}
     {|O_s^c(T)|}, \\
r_{f\rightarrow o}(T)=
\frac{\left|\{p\in F_s^c(T):\ell_t(Tp)=\mathrm{occ}\}\right|}
     {|F_s^c(T)|}.
\end{align}
The first ratio measures source obstacles that the stored map identifies as free space. The second measures source free space that the stored map identifies as occupied.

For trajectory coverage, we divide the recent source trajectory into four consecutive bins of equal traveled distance (i.e., $2\,\mathrm{m}$ each for a full $8\,\mathrm{m}$ window). Each occupied-cell match is assigned to the bin containing the robot pose from which that cell was observed. A bin is supported when it contributes at least ten matches. This test prevents a candidate from being accepted because of agreement with only one short portion of the trajectory.
\begin{enumerate}[leftmargin=*,label=\arabic*.]
    \item \textbf{Minimum evidence:} both source and target contain at least 120 occupied points.
    \item \textbf{Occupancy consistency:} at least 80 occupied and 120 free source cells must be comparable. Among them, $r_{o\rightarrow f}(T)\leq0.15$ and $r_{f\rightarrow o}(T)\leq0.10$. Moreover, at least 70\% of the comparable occupied cells must lie within $0.15\,\mathrm{m}$ of a target obstacle, and these matches must cover at least 50 distinct target cells. These conditions jointly require sufficient overlap, few occupied/free contradictions, and positive obstacle agreement.
    \item \textbf{Trajectory coverage:} at least three of the four trajectory bins must be supported.
    \item \textbf{Refinement quality:} ICP fitness is at least 0.10, RMSE is at most $0.20\,\mathrm{m}$, and the refined alignment contains at least 50 inliers.
    \item \textbf{Prior consistency:} the current robot pose, after applying $T$, lies inside the agent-reported target region.
    \item \textbf{Ambiguity margin:} the geometric score exceeds that of the best spatially distinct valid alternative by at least 0.03.
\end{enumerate}
The geometric score rewards occupied-cell agreement and spatially distributed target support while penalizing occupied-to-free and free-to-occupied conflicts. Failure of any hard-conflict limit, positive-evidence requirement, refinement diagnostic, region constraint, or ambiguity margin rejects the prior. In that case, mapping continues in the current local frame and the stored global frame is left unchanged.

\paragraph{Frame handoff and correction.}
After acceptance, the estimated $\mathrm{SE}(2)$ transform is applied through a Cartographer trajectory handoff. The system rebases the current odometry frame at the scan sequence used for matching and initializes a new trajectory against the frozen exploration map. A corrective relocalization report made after an earlier localization passes through the same acceptance gates. The previous trajectory remains active until the new anchor is confirmed, allowing the system to roll back if the correction fails.

\subsection{Geometric Navigation}
\label{sec:supp-navigation}

Given a memory node or unvisited-area marker, the executor retrieves its map position and produces an executable action sequence. In realworld deployment on Unitree Go2, we use the path planner implementation in dimos. In simulator, the path planner follows three stages: cost-map construction, A* search, and path discretization. The stages deliberately separate safety constraints from global route search and robot-specific action compilation.

\paragraph{Cost-map construction.}
We first convert the planning occupancy grid into a cost map that distinguishes hard safety constraints from soft path preferences. Free cells are initially traversable at zero cost. Occupied cells are blocked, and unknown cells are also blocked by default. If unknown traversal is explicitly enabled, unknown cells become traversable with a cost of $c_{\mathrm{unknown}}=80$, making known free space preferable without making exploration impossible.

Next, we inflate every blocked cell by
\begin{equation}
n_{\mathrm{inf}}=\left\lceil\frac{r_{\mathrm{inf}}}{\rho}\right\rceil
\end{equation}
grid cells, where the map resolution is $\rho=0.05\,\mathrm{m}$ and the inflation radius is $r_{\mathrm{inf}}=0.15\,\mathrm{m}$. Inflation reserves clearance for the robot footprint instead of treating the robot as a point. A disk of radius $r_{\mathrm{clear}}=0.20\,\mathrm{m}$ around the current pose is then cleared. This local exception prevents a small mapping or pose error from declaring the robot's own starting cell invalid.

Finally, we compute the distance $D(p)$ from each traversable cell $p$ to the inflated obstacle set and add the soft penalty
\begin{equation}
c_{\mathrm{clear}}(p)=\max\!\left(0,c_{\mathrm{grad}}-3D(p)\right),
\qquad c_{\mathrm{grad}}=18.
\end{equation}
Cells near obstacles therefore remain traversable when they satisfy the hard-clearance constraint, but routes through better-cleared space receive lower cost. The output is a finite cost for every traversable cell and infinite cost for every blocked cell.

\paragraph{A* route search.}
The planner first converts the metric start and goal into grid coordinates. Mapping noise may place either coordinate just inside an inflated cell, so each is snapped to the nearest traversable cell within $d_{\mathrm{snap}}=1.0\,\mathrm{m}$. If no such cell exists, planning terminates with a distinct start- or goal-not-traversable status rather than silently selecting a distant substitute.

We then run A* \cite{hart1968formal} over the eight-connected grid with octile distance as the heuristic. A cardinal move has base length 1 and a diagonal move has base length $\sqrt{2}$. Entering cell $v$ incurs
\begin{equation}
c(u,v)=b(u,v)\left(1+\frac{C(v)}{100}\right),
\end{equation}
where $b(u,v)\in\{1,\sqrt{2}\}$ is the move length and $C(v)$ is the destination-cell cost. This preserves geometric path length while biasing search away from unknown or poorly cleared cells. A diagonal transition is rejected if either adjacent cardinal cell is blocked, preventing a path from cutting through an obstacle corner. When the goal is reached, parent links are reconstructed and converted back into metric SLAM coordinates. If the queue is exhausted first, the planner returns an explicit unreachable status.

\paragraph{Path simplification and action compilation.}
The raw grid path contains many adjacent cells and cannot be directly executed by a robot with quantized actions. We convert it into actions in three stages.
\begin{enumerate}[leftmargin=*,label=\arabic*.]
    \item \textbf{Line-of-sight simplification.} Starting from the first path cell, we find the farthest later cell connected to the current anchor by a collision-free Bresenham line. That cell becomes the next waypoint and the new anchor. Repeating this process removes redundant cells while preserving collision-free straight segments. The retained waypoints are then converted into metric SLAM coordinates, and the first waypoint is removed because it represents the robot's current position.
    \item \textbf{Quantized action selection.} For each waypoint, we compute the ideal heading and round it to an initial number of $10^\circ$ turns. Because independent rounding can make the subsequent $0.25\,\mathrm{m}$ forward step unsafe or unproductive, we test nearby integer turn counts in order of remaining heading error. Each candidate is simulated before execution. We reject it if the residual heading error is excessive, its swept forward segment intersects a blocked cell, or the step increases the waypoint distance by more than one map cell. The first valid candidate contributes its individual left/right turns followed by one forward action.
    \item \textbf{Validation and recovery.} Every proposed forward sweep is sampled at intervals no larger than $0.025\,\mathrm{m}$. If no candidate is valid, line-of-sight waypoints are recomputed from the current pose. The compiler reports a blocked-sweep failure if the same segment remains invalid. Execution terminates successfully once the robot lies within $r_{\mathrm{goal}}=0.45\,\mathrm{m}$ of the target. Reaching the end of the compiled route outside this radius yields a partial-route status. During online execution, a collision, blocked motion, or disagreement between the predicted and updated SLAM poses triggers global replanning, with at most two replanning attempts by default.
\end{enumerate}
This simulate-before-commit procedure preserves the intent of the continuous grid path under a discrete action space and exposes explicit failure states instead of issuing an unsafe forward motion.

%% file: tables/eqa_type.tex
\begin{table}[t]
\caption{\textbf{EQA task distribution by answer type and question type.} $N$ is the number of tasks in each group. Each panel sums to 150.}
\label{tab:eqa-breakdown}
\centering
\footnotesize
\begin{minipage}[t]{0.48\linewidth}
\centering
\textbf{(a) Answer type}\\[3pt]
\setlength{\tabcolsep}{3pt}
\begin{tabular*}{\linewidth}{@{\extracolsep{\fill}}lr@{}}
\toprule
Type & $N$ \\
\midrule
Integer & 31 \\
Multi-choice & 12 \\
Ordered list & 8 \\
Single choice & 76 \\
Unordered set & 5 \\
Yes/no & 18 \\
\bottomrule
\end{tabular*}
\end{minipage}
\hfill
\begin{minipage}[t]{0.48\linewidth}
\centering
\textbf{(b) Question type}\\[3pt]
\setlength{\tabcolsep}{3pt}
\begin{tabular*}{\linewidth}{@{\extracolsep{\fill}}lr@{}}
\toprule
Type & $N$ \\
\midrule
Attribute & 41 \\
Counting & 34 \\
Existence & 10 \\
Functional reasoning & 14 \\
Location & 12 \\
Relation & 39 \\
\bottomrule
\end{tabular*}
\end{minipage}
\end{table}

\begin{table}[t]
\caption{\textbf{Embodied question answering on 150 Habitat-GS tasks.} Accuracy is the percentage of questions answered exactly correctly. Except for Uni-NaVid, all methods use \texttt{GPT-6.1-sol} as the reasoning model.}
\label{tab:eqa-overall}
\centering
\footnotesize
\setlength{\tabcolsep}{4pt}
\begin{tabular*}{\linewidth}{@{\extracolsep{\fill}}lcc@{}}
\toprule
Method & Model & Acc. (\%)$\uparrow$ \\
\midrule
SuperNav~\cite{zhang2026supernav} & GPT-6.1-sol & 81.33 \\
FAST-EQA~\cite{zhang2026fasteqa} & GPT-6.1-sol & 55.33 \\
GraphEQA~\cite{saxena2024grapheqa} & GPT-6.1-sol & 28.00 \\
3D-Mem~\cite{yang2025threedmem} & GPT-6.1-sol & 32.67 \\
Uni-NaVid~\cite{zhang2025uninavid} & Uni-NaVid-7B & 22.00 \\
\midrule
\textbf{Ours} & GPT-6.1-sol & \textbf{87.33} \\
\bottomrule
\end{tabular*}
\end{table}